\documentclass[runningheads]{llncs}
\usepackage[T1]{fontenc}
\usepackage{graphicx}
\usepackage{hyperref}
\usepackage{booktabs}
\usepackage{multirow}
\usepackage{eqparbox}
\usepackage{amsmath}
\usepackage{subcaption}
\usepackage{float}
\usepackage{cite}
\usepackage{tikz}
\usetikzlibrary{arrows.meta}
\usepackage{color}

\begin{document}
\title{ConRTF: Edge-Constrained Boundary Distribution Refinement for Realtime TransFormer Table Structure Recognition}
\titlerunning{ConRTF: Edge-Constrained Refinement for Table Structure Recognition}
%
\author{Eliott Thomas\inst{1,2}\orcidID{0009-0008-5266-8797}\textsuperscript{\,$\star$} \and
Tri-Cong Pham\inst{1}\orcidID{0000-0002-5507-6454}\textsuperscript{\,$\star$} \and
Micka\"el Coustaty\inst{1}\orcidID{0000-0002-0123-439X} \and
Aur\'elie Joseph\inst{2}\orcidID{0000-0002-5499-6355} \and
Gaspar Deloin\inst{2}\orcidID{0009-0007-2449-5385} \and
Vincent Poulain d'Andecy\inst{2}\orcidID{0009-0008-5515-3561} \and
Jean-Marc Ogier\inst{1}\orcidID{0000-0002-5666-475X} \and
Antoine Doucet\inst{1,3}\orcidID{0000-0001-6160-3356}}
\authorrunning{E. Thomas et al.}
\institute{L3i, La Rochelle University, France\\
\email{\{eliott.thomas, cong.pham, mickael.coustaty, jean-marc.ogier, antoine.doucet\}@univ-lr.fr} \and
Yooz, La Rochelle, France\\
\email{\{aurelie.joseph, gaspar.deloin, vincent.poulaindandecy\}@getyooz.com} \and
Faculty of Computer and Information Science, University of Ljubljana, Slovenia}
\maketitle              
\renewcommand{\thefootnote}{}\footnotetext{$\star$~Equal contribution.}\renewcommand{\thefootnote}{\arabic{footnote}}
\begin{abstract}
\vspace{-3.5mm}
Table Structure Recognition (TSR) aims to recover the row and column layout of tables from document images, a key step in document understanding pipelines. Accurate TSR depends on precise boundary localization:
  small errors in row or column boundaries can propagate into incorrect cell assignments and structural inconsistencies. Yet detection-based approaches treat table elements as generic objects, ignoring a
  fundamental property of table layout: rows and columns play structurally distinct roles and their boundaries carry unequal importance. We propose an Edge-constrained Fine-grained Localization loss (EFL) that formalizes this structural asymmetry by encoding table-specific geometric priors into the training objective:
  row-like elements are supervised with emphasis on their horizontal boundaries, while column-like elements prioritize vertical boundaries. Implemented within a real-time detector with distribution-based boundary refinement (D-FINE), EFL operates during training only and guides boundary refinement toward
  structurally meaningful adjustments with no change to the inference pipeline. The proposed approach, ConRTF, is also data-efficient, maintaining robust accuracy with as few as 2k--3k annotated tables. Experiments
   on PubTables-1M and two private datasets show consistent improvements over the optimized baseline and several real-time detectors including RT-DETRv2 and YOLOv10-11, with gains of up to +1.6 GriTS points at equal inference
  speed.
\vspace{-2mm}
\end{abstract}

\section{Introduction}
Tables organize information across scientific articles, financial reports, forms, and business documents. Table Structure Recognition (TSR) recovers this layout from document images and is central to document understanding~\cite{IJDAR_Zanibbi_Survey_2004,ACM_Kasem_Survey_2024}.

\begin{figure}[t]
\centering
\begin{tikzpicture}[>=Stealth, font=\small, scale=0.8, every node/.style={scale=0.8}]


\node[anchor=south, font=\bfseries\large] at (3, 3.2) {Row-like element};
\node[anchor=south, font=\normalsize, gray] at (3, 2.8) {(rows, headers, sections)};

\fill[red!8] (0,-0.75) rectangle (6,0.75);
\draw[red, line width=3pt] (0, 0.75) -- (6, 0.75);
\draw[red, line width=3pt] (0, -0.75) -- (6, -0.75);
\draw[gray, line width=1pt, dashed] (0, -0.75) -- (0, 0.75);
\draw[gray, line width=1pt, dashed] (6, -0.75) -- (6, 0.75);

\draw[red, {Stealth}-] (3, 1.05) -- (3, 1.75) node[above, font=\normalsize] {$\lambda_h$};
\draw[red, {Stealth}-] (3, -1.05) -- (3, -1.75) node[below, font=\normalsize] {$\lambda_h$};
\draw[gray, {Stealth}-] (-0.3, 0) -- (-0.8, 0) node[left, font=\normalsize] {$\lambda_l$};
\draw[gray, {Stealth}-] (6.3, 0) -- (6.8, 0) node[right, font=\normalsize] {$\lambda_l$};

\node[red, font=\footnotesize\itshape, rotate=0] at (3, 0.35) {defining edges};
\node[gray, font=\footnotesize\itshape, rotate=90] at (0.5, 0) {extent};

\node[anchor=south, font=\bfseries\large] at (12, 3.2) {Column-like element};
\node[anchor=south, font=\normalsize, gray] at (12, 2.8) {(columns)};

\fill[green!8] (10,-1.5) rectangle (14,1.5);
\draw[gray, line width=1pt, dashed] (10, 1.5) -- (14, 1.5);
\draw[gray, line width=1pt, dashed] (10, -1.5) -- (14, -1.5);
\draw[green!60!black, line width=3pt] (10, -1.5) -- (10, 1.5);
\draw[green!60!black, line width=3pt] (14, -1.5) -- (14, 1.5);

\draw[green!60!black, {Stealth}-] (9.7, 0.0) -- (9.0, 0.0) node[left, font=\normalsize] {$\lambda_h$};
\draw[green!60!black, {Stealth}-] (14.3, 0.0) -- (15.0, 0.0) node[right, font=\normalsize] {$\lambda_h$};
\draw[gray, {Stealth}-] (12, 1.8) -- (12, 2.3) node[above, font=\normalsize] {$\lambda_l$};
\draw[gray, {Stealth}-] (12, -1.8) -- (12, -2.3) node[below, font=\normalsize] {$\lambda_l$};

\node[green!60!black, font=\footnotesize\itshape, rotate=90] at (10.5, 0) {defining edges};
\node[gray, font=\footnotesize\itshape, rotate=0] at (12, 1.15) {extent};

\end{tikzpicture}
\caption{EFL edge weighting. Structurally defining boundaries (thick, colored) receive weight $\lambda_h$ (high), while extent boundaries (dashed, gray) receive $\lambda_l$ (low), with $\lambda_l < \lambda_h$. The asymmetry encodes the geometric prior that horizontal edges matter more for rows and vertical edges matter more for columns.}
\label{fig:efl_concept}
\end{figure}
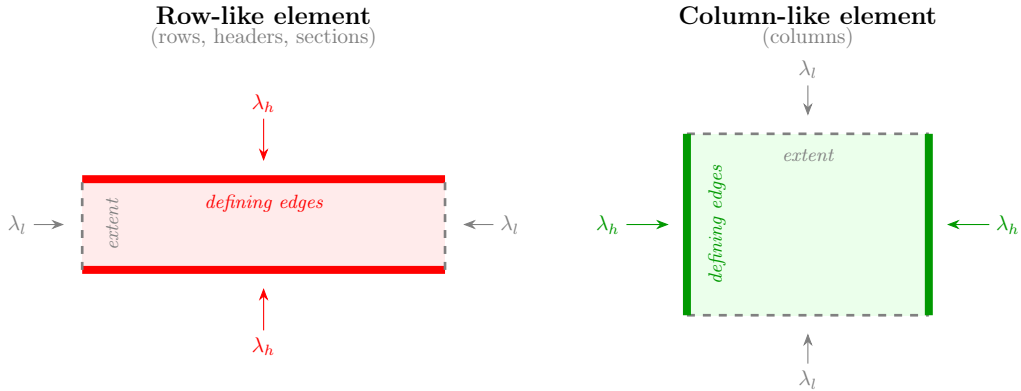

TSR~\cite{CVPR_Smock_PubTables1M_2022,ICDAR_ThomasQUEST_2025,WACV_Eliott_RAPTOR_2025,AX_Xiao_TableCenterNet_2025} remains challenging because of its sensitivity to localization errors. Unlike generic object detection, small inaccuracies in row or column boundaries propagate into structural inconsistencies. Structure-aware metrics such as TEDS~\cite{ECCV_Zhong_TEDS_2020} and GriTS~\cite{ICDAR_Smock_GriTS_2023} penalize even minor boundary misalignments, so accurate TSR demands precise boundary localization rather than coarse object-level detection.
Recent approaches leverage Transformer-based architectures and multi-stage pipelines~\cite{ICDAR_Hou_TABLET_2025,EMNLP_Zhang_UniTabNet_2024,IJCAI_KhangTFLOP_2024,PR_LONG_LOREPlus_2025} but their depth and post-processing hinder real-time deployment. Many also treat table elements via point-based or generic detection~\cite{ICDAR_Smock_TATR_2023,ESA_Xiao_CascadeTSRDet_2025}, where boundary uncertainty and table-specific geometry are only implicitly handled. Boundary-aware modeling for joint localization and efficiency remains underexplored.
We revisit TSR from a boundary-centric perspective. We model table elements as row-like and column-like objects with distribution-based boundary representations. This matches the elongated geometry of rows, columns, headers, and section separators. Distribution-based boundaries enable fine-grained refinement within a compact real-time architecture.
We introduce an Edge-constrained Fine-grained Localization loss (EFL) that aligns refinement with table layout semantics: horizontal boundaries are emphasized for row-like elements, vertical ones for column-like elements (Figure~\ref{fig:efl_concept}). EFL guides refinement toward structurally meaningful adjustments at no inference cost.

We also address practical settings where large-scale annotations are scarce. Combining boundary distribution refinement with self-distillation, ConRTF maintains accuracy with small annotated tables.

The main contributions are:

\begin{itemize}
    \item We reformulate TSR as a boundary-centric task with distribution-based boundary representations, enabling fine-grained localization and accurate structure recovery within a real-time architecture.

    \item We introduce EFL, which encodes table-specific geometric priors into boundary supervision via edge weights derived from an IoU-sensitivity analysis. EFL operates during training only, adding no inference cost.

    \item We demonstrate that the proposed approach is data-efficient, maintaining robust structural accuracy even when trained with only 2k--3k annotated tables, making it suitable for deployment under limited annotation budgets.

    \item We conduct extensive experiments on the large-scale PubTables-1M benchmark and two private datasets, evaluating performance using TEDS and GriTS metrics together with detailed runtime analyses, validating both structural accuracy and real-time efficiency.
    

\end{itemize}

\section{Related Work}
\textbf{Table Structure Recognition.}
Table structure recognition (TSR) aims to recover the row-column layout of a table from its image. Early systems relied on ruling lines, whitespace analysis, and hand-crafted heuristics~\cite{IJDAR_Zanibbi_Survey_2004}. Deep learning methods have since replaced these pipelines, organized into three main families~\cite{ACM_Kasem_Survey_2024}. Detection-based methods treat rows, columns, and cells as objects, from Faster R-CNN~\cite{ICDAR_Schreiber_DeepDeSRT_2017} to DETR-style architectures on PubTables-1M~\cite{ICDAR_Smock_TATR_2023,CVPR_Smock_PubTables1M_2022} and cascade detectors with tailored anchor ratios~\cite{ESA_Xiao_CascadeTSRDet_2025}. Sequence-based methods generate structure as token sequences, using HTML prediction~\cite{CVPR_Nassar_TableFormer_2022}, compact tokenizations~\cite{ICDAR_Maksym_TSR_2023}, or pointer mechanisms~\cite{IJCAI_KhangTFLOP_2024}. Regression-based methods directly predict cell coordinates with self-supervised pre-training~\cite{PR_LONG_LOREPlus_2025}. Despite this diversity, none explicitly encodes the geometric asymmetry between row-like and column-like elements into the training signal.

\textbf{Real-time Object Detection.}
DETR~\cite{ECCV_Carion_DETR_2020} reformulated object detection as set prediction with a transformer, eliminating hand-crafted components such as anchors and non-maximum suppression. Deformable DETR~\cite{ICLR_Zhu_DeformableDETR_2021} replaced dense attention with sparse deformable sampling, reducing convergence time by an order of magnitude. DINO~\cite{ICLR_Zhang_DINO_2023} further improved accuracy through contrastive denoising training and mixed query selection. RT-DETR~\cite{CVPR_Zhao_RT-DETR_2024} introduced a hybrid encoder combining attention-based and CNN-based feature fusion, achieving real-time inference while matching DETR accuracy. D-FINE~\cite{ICLR_Peng_DFINE_2025} builds on GFocal's distribution-based regression~\cite{NeurIPS_Xiang_GFocal_2020,CVPR_Xiang_GFocalv2_2021} with iterative refinement (FDR) and self-distillation (GO-LSD), setting a new accuracy-speed trade-off. In parallel, the YOLO family~\cite{AIR_Sapkota_YOLO_Survey_2025} has continued to push single-stage detection speed, with recent versions~\cite{NeurIPS_Wang_YOLOv10_2024,Jocher_YOLO11_2024,NeurIPS_Tian_YOLOv12_2025} approaching transformer-based accuracy. We build on D-FINE because its iterative refinement of distribution-based boundaries provides a natural entry point for encoding table-specific geometric priors.

\textbf{Fine-grained Localization.}
Standard detectors regress bounding box coordinates directly, which can be imprecise for elongated or off-centered objects. Distribution-based methods~\cite{NeurIPS_Xiang_GFocal_2020,CVPR_Xiang_GFocalv2_2021} model each boundary as a probability distribution over candidate offsets, improving robustness. D-FINE~\cite{ICLR_Peng_DFINE_2025} extends this idea with iterative refinement and global self-distillation. However, these approaches treat all four edges symmetrically, regardless of geometry or semantic role.

\textbf{Geometry-aware Loss Design.}
Incorporating geometric priors into loss functions has been explored in oriented object detection for aerial imagery. Yang et al.~\cite{ICML_Yang_GWD_2021} model rotated boxes as Gaussian distributions with aspect-ratio-adaptive gradients, extended via Kullback--Leibler (KL) divergence~\cite{NeurIPS_Yang_KLD_2021}. ARS-DETR~\cite{TGRS_Zeng_ARSDETR_2024} explicitly introduces aspect-ratio-sensitive loss weighting for elongated objects. However, these methods target rotated bounding boxes. Table elements are axis-aligned but exhibit a different asymmetry: a row's horizontal edges define inter-row boundaries critical for structure, while its vertical edges merely mark table extent. EFL addresses this edge asymmetry within a distribution-based refinement method.

\section{Proposed Method}

This section presents the proposed approach for real-time table structure recognition. We first reformulate TSR from a boundary-centric perspective, modeling table elements as row-like and column-like objects and representing their boundaries using distribution-based modeling (Section~\ref{sec:preliminaries}). We then introduce an Edge-constrained Fine-grained Localization loss that encodes table-specific geometric priors into boundary supervision (Section~\ref{sec:efl}). Section~\ref{sec:system} describes the complete ConRTF system.

\begin{figure*}[!t]
	\centering
	\includegraphics[width=1.0\textwidth]{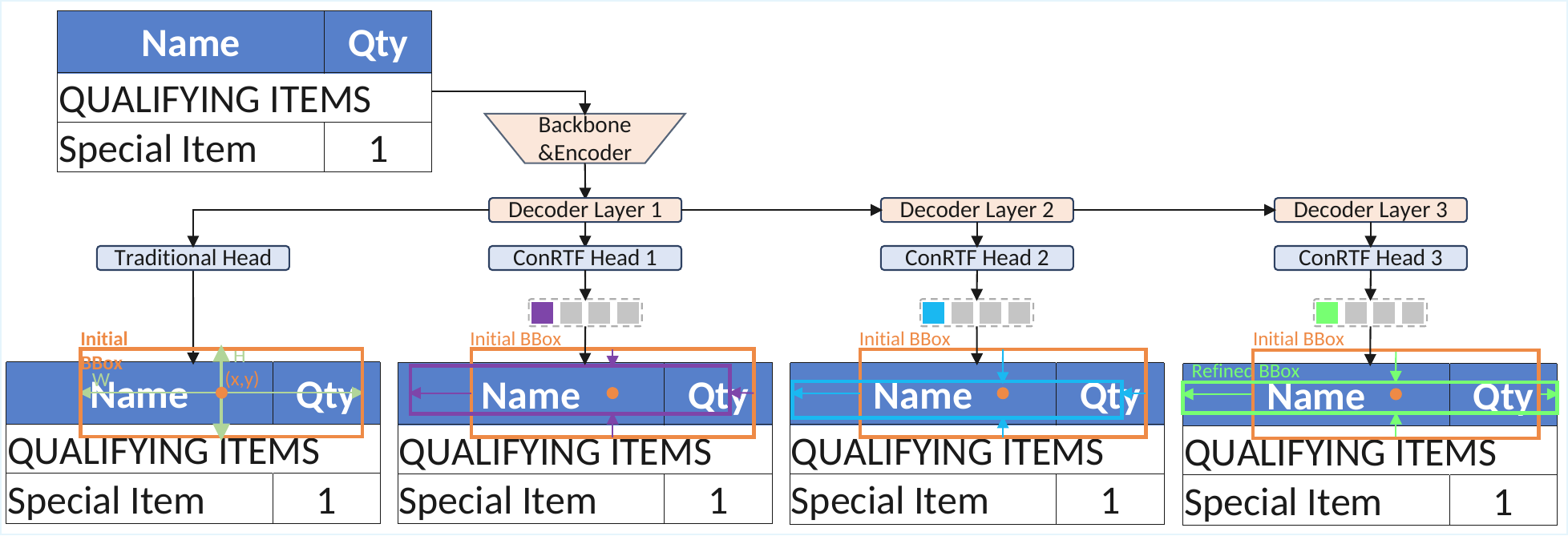}
	\caption{Overview of ConRTF. Each decoder layer iteratively refines boundary distributions in a residual manner through non-uniform weighting functions. During training, EFL (Figure~\ref{fig:efl_concept}) applies class-dependent edge weights to guide refinement toward structurally defining boundaries.}
	\label{fig-ConRTF}
\end{figure*}

\subsection{Preliminaries: Distribution-based Boundary Refinement}
\label{sec:preliminaries}

We formulate TSR as the task of accurately localizing table element boundaries rather than detecting cells or generating structural sequences. This is motivated by the observation that most structural errors originate from imprecise row or column boundaries, where small localization deviations propagate into incorrect adjacency relationships. To enable robust localization, we represent each boundary as a probability distribution over candidate offsets. GFocal~\cite{NeurIPS_Xiang_GFocal_2020,CVPR_Xiang_GFocalv2_2021} introduced this distribution-based regression, and D-FINE~\cite{ICLR_Peng_DFINE_2025} extended it with iterative refinement and a non-uniform weighting function within a real-time DETR architecture~\cite{CVPR_Zhao_RT-DETR_2024,ECCV_Carion_DETR_2020}. We adopt D-FINE as our base detector and review the key components. The first decoder layer predicts an initial bounding box $\mathbf{b}^{0}=\{x, y, W, H\}$, where $\{x, y\}$ denotes the box center and $\{W, H\}$ the width and height. The center $\mathbf{c}^{0}=\{x, y\}$ and edge distances $\mathbf{d}^{0}=\{t, b, l, r\}$ from the center to the top, bottom, left, and right edges are derived accordingly. At the $l$-th decoder layer, boundary offset representations $\mathbf{P}^{l}=\{P_t^{l}, P_b^{l}, P_l^{l}, P_r^{l}\}$ encode residual offsets with respect to $\mathbf{d}^{0}$, as illustrated in Figure~\ref{fig-ConRTF}. The residual edge offsets are computed through a scale-aware transformation:
\begin{equation}
    \Delta \mathbf{d}^{l} = \{H, H, W, W\} \cdot f(\mathbf{P}^{l}),
    \label{eq:edge_refinement}
\end{equation}
where $f(\cdot)$ denotes a distribution-to-offset transformation defined below.

Distributions are refined across decoder layers through residual logit accumulation rather than prediction from scratch:
\begin{equation}
    P^l(n) = \mathrm{Softmax}\!\left(\Delta \mathrm{logits}^l(n) + \mathrm{logits}^{l-1}(n)\right),
    \label{eq:residual}
\end{equation}
where each layer predicts incremental logits $\Delta \mathrm{logits}^l(n)$ that correct the previous layer's estimate, enabling coarse-to-fine boundary refinement.

$f(\cdot)$ in Eq.~\eqref{eq:edge_refinement} uses a non-uniform weighting function~\cite{CVPR_Xiang_GFocalv2_2021}:
\begin{equation}
    f\!\left(P^{l}\right) = \sum_{n=0}^{N} W(n)\, P^{l}_{n},
    \label{eq:f_transform}
\end{equation}
\begin{equation}
    W(n) =
    \begin{cases}
    -2a, & n = 0, \\[4pt]
    c - c\!\left(\dfrac{a}{c}+1\right)^{\frac{N-2n}{N-2}}, & 1 \le n < \dfrac{N}{2}, \\[10pt]
    - c + c\!\left(\dfrac{a}{c}+1\right)^{\frac{-N+2n}{N-2}}, & \dfrac{N}{2} \le n \le N-1, \\[10pt]
    2a, & n = N,
    \end{cases}
    \label{eq:weighting}
\end{equation}
where $n$ indexes the discrete spatial bins, $N$ is the total number of bins, and $a$ and $c$ are hyper-parameters controlling the magnitude and curvature of the weighting function. The shape of $W(n)$ is adaptive: near the center of the bin range ($n \approx N/2$), $W(n)$ varies smoothly, enabling fine-grained adjustments for near-accurate predictions. Near the extremes, the curvature increases, providing flexibility for larger corrections.

D-FINE supervises boundary distributions with a Fine-Grained Localization (FGL) loss that encourages predicted distributions to concentrate on the ground-truth offset. For each of $K$ matched prediction-ground-truth pairs (obtained via Hungarian matching~\cite{ECCV_Carion_DETR_2020}), the ground-truth relative offset $\phi = (\mathbf{d}^{\mathrm{GT}} - \mathbf{d}^{0}) / \{H, H, W, W\}$ generally falls between two adjacent bins $n_{\leftarrow}$ and $n_{\rightarrow}$ in the warped space defined by $W$. The total FGL loss sums over all $L$ decoder layers, $\mathcal{L}_{\mathrm{FGL}} = \sum_{l=1}^{L} \mathcal{L}_{\mathrm{FGL}}^l$, where:
\begin{equation}
    \mathcal{L}_{\mathrm{FGL}}^l = \sum_{k=1}^{K} \mathrm{IoU}_k \cdot \left( \omega_{\leftarrow} \cdot \mathrm{CE}(P^l(n)_k, n_{\leftarrow}) + \omega_{\rightarrow} \cdot \mathrm{CE}(P^l(n)_k, n_{\rightarrow}) \right)
    \label{eq:fgl}
\end{equation}
\begin{equation}
    \omega_{\leftarrow} = \frac{|\phi - W(n_{\rightarrow})|}{|W(n_{\leftarrow}) - W(n_{\rightarrow})|}, \quad \omega_{\rightarrow} = \frac{|\phi - W(n_{\leftarrow})|}{|W(n_{\leftarrow}) - W(n_{\rightarrow})|}
    \label{eq:interp_fgl}
\end{equation}
where $\mathrm{IoU}_k$ is the IoU between prediction $k$ and its matched ground truth and $\mathrm{CE}$ denotes cross-entropy. Each edge contributes to the loss with equal weight.

D-FINE additionally employs Global Optimal Localization Self-Distillation (GO-LSD)~\cite{ICLR_Peng_DFINE_2025}, which transfers refined boundary distributions from the final decoder layer to shallower layers through KL-divergence-based distillation. This stabilizes early-layer predictions and improves convergence, particularly under limited training data. We adopt GO-LSD unchanged from D-FINE.

\subsection{Edge-constrained Fine-grained Localization Loss}
\label{sec:efl}

The FGL loss defined in Eq.~\eqref{eq:fgl} supervises all four edges of each element with equal weight, which is appropriate for generic object detection where no edge direction is inherently more important. However, table elements exhibit a structural asymmetry that this uniform treatment ignores: row-like elements (rows, headers, section separators) are defined by their horizontal boundaries, while columns are defined by their vertical boundaries. Misaligning a row's top or bottom edge corrupts the table structure more than a comparable error on its lateral edges, and vice versa for columns. Inspired by FGL, we propose an Edge-constrained Fine-grained Localization (EFL) loss that re-weights boundary supervision according to each element's structural role. The total EFL loss sums over all $L$ decoder layers, $\mathcal{L}_{\mathrm{EFL}} = \sum_{l=1}^{L} \mathcal{L}_{\mathrm{EFL}}^l$, where:
\begin{equation}
    \mathcal{L}_{\mathrm{EFL}}^l = \sum_{k=1}^{K} \mathrm{IoU}_k \cdot \left( \boldsymbol{\omega}^*_{\leftarrow} \cdot \mathrm{CE}(P^l(n)_k, n_{\leftarrow}) + \boldsymbol{\omega}^*_{\rightarrow} \cdot \mathrm{CE}(P^l(n)_k, n_{\rightarrow}) \right)
    \label{eq:efl}
\end{equation}

Compared to FGL (Eq.~\eqref{eq:interp_fgl}), EFL modifies the interpolation weights $\boldsymbol{\omega}^*$ by introducing a class-dependent edge weight:
\begin{equation}
    \boldsymbol{\omega}^*_{\leftarrow} = \frac{|\phi - W(n_{\rightarrow})|}{|W(n_{\leftarrow}) - W(n_{\rightarrow})|} \cdot e_w, \quad \boldsymbol{\omega}^*_{\rightarrow} = \frac{|\phi - W(n_{\leftarrow})|}{|W(n_{\leftarrow}) - W(n_{\rightarrow})|}  \cdot e_w
    \label{eq:interp_efl}
\end{equation}

The loss in Eqs.~\eqref{eq:efl}--\eqref{eq:interp_efl} is computed independently for each of the four edges, and $e_w$ selects the corresponding scalar weight per edge direction. The edge weight vector $e_w \in \{e_w^r, e_w^c\}$ is assigned based on the element class:
\begin{equation}
    e_w^r = [\lambda_{h}, \lambda_{h}, \lambda_{l}, \lambda_{l}], \quad e_w^c = [\lambda_{l}, \lambda_{l}, \lambda_{h}, \lambda_{h}]
    \label{eq:edge_weights}
\end{equation}
where the four components correspond to the top, bottom, left, and right edges, respectively, and $\lambda_h > \lambda_l$ (Figure~\ref{fig:efl_concept}). Row-like elements use $e_w^r$, receiving stronger supervision on their structurally defining horizontal boundaries, while columns use $e_w^c$, prioritizing vertical boundaries. EFL is applied during training only and introduces no additional cost at inference.

The values of $\lambda_h$ and $\lambda_l$ can be derived from an IoU-sensitivity argument. For any table element, let $S$ and $L$ denote the short and long dimensions, respectively, with aspect ratio $\mathrm{AR} = L/S > 1$. Defining edges run along $L$ and bound $S$, while extent edges run along $S$ and bound $L$. For rows, $S$ is the height and $L$ the width. For columns, the roles are swapped. The derivation is identical in both cases.

Consider an element with ground-truth dimensions $L \times S$. If one defining edge is shifted outward by $\varepsilon$ pixels, the predicted box becomes $L \times (S + \varepsilon)$. Since the ground truth fits inside the prediction,
\begin{equation}
    \mathrm{IoU}_{\mathrm{def}}
      = \frac{S}{S + \varepsilon}
      = \frac{1}{1 + \varepsilon/S}
      \approx 1 - \frac{\varepsilon}{S}.
    \label{eq:iou_def}
\end{equation}
The same perturbation on an extent edge gives $\mathrm{IoU}_{\mathrm{ext}} \approx 1 - \varepsilon/L$. Since $S \ll L$, defining-edge errors cause $\mathrm{AR}$ times more IoU degradation than extent-edge errors.

We model expected localization error on an edge as inversely proportional to its loss weight: defining edges (weight $\lambda_h$) have expected error $C/\lambda_h$, extent edges (weight $\lambda_l$) have expected error $C/\lambda_l$. Each element has two defining and two extent edges. Substituting into the IoU drops and summing (valid to first order for small perturbations), the total IoU degradation is proportional to
\begin{equation}
    f(\lambda_h) = \frac{1}{\lambda_h \cdot S}
                  + \frac{1}{\lambda_l \cdot L}.
    \label{eq:iou_total}
\end{equation}
We minimize $f$ subject to a budget constraint: the total supervision weight must equal FGL's uniform total ($2\lambda_h + 2\lambda_l = 4$, i.e., \ $\lambda_h + \lambda_l = 2$). Substituting $\lambda_l = 2 - \lambda_h$, differentiating, and setting to zero gives
\begin{equation}
    \frac{1}{(2 - \lambda_h)^2 \cdot L} = \frac{1}{\lambda_h^2 \cdot S}.
    \label{eq:lambda_deriv}
\end{equation}
At the optimum, the marginal cost of shifting weight toward defining edges exactly equals the marginal cost on extent edges. Solving:
\begin{equation}
    \lambda_h^{*} = \frac{2\sqrt{\mathrm{AR}}}{1 + \sqrt{\mathrm{AR}}},
    \qquad
    \lambda_l^{*} = \frac{2}{1 + \sqrt{\mathrm{AR}}}.
    \label{eq:lambda_opt}
\end{equation}
When $\mathrm{AR} = 1$ (square elements), this reduces to $\lambda_h = \lambda_l = 1$, recovering the uniform FGL loss. Since EFL uses a single $(\lambda_h, \lambda_l)$ pair across all element classes, the shared optimum is constrained by the least elongated class. Setting $\lambda_h$ at the optimum of a highly elongated class would reduce $\lambda_l$ to a level that degrades defining-edge supervision for less elongated classes. The concrete values are determined by the dataset aspect ratios and reported in Section~\ref{sec:impl}.

\subsection{Architectural Design and Optimization}
\label{sec:system}

ConRTF adopts the D-FINE architecture~\cite{ICLR_Peng_DFINE_2025} with an HGNetv2-X backbone encoder and a Transformer decoder with $L$ refinement layers. The total training loss follows D-FINE, combining classification, box regression, and boundary distribution supervision. The sole modification is the replacement of FGL (Eq.~\eqref{eq:fgl}) with EFL (Eq.~\eqref{eq:efl}). GO-LSD self-distillation is applied unchanged.

At inference, boundary distributions are decoded into bounding boxes via Eq.~\eqref{eq:edge_refinement}, producing detections for all element classes. Table cells are reconstructed by computing row-column intersections following TATR~\cite{ICDAR_Smock_TATR_2023}. The inference pipeline is identical to D-FINE, incurring no additional cost from EFL.

\begin{table}[ht]
	\centering
	\caption{Datasets.}
	\label{tab:datasets}
	\setlength{\tabcolsep}{8pt}
	\begin{tabular}{clrrrr}
		\toprule
		\# & Dataset  &  Train & Validation & Test & Total \\
		\midrule
		1 & PubTables-1M & 701,275 & 87,835 & 86,917 & \textbf{876,027} \\
		2 & Business & 3,054 & 654 & 654 & \textbf{4,362} \\
		3 & Finance & 2,123 & 374 & 272 & \textbf{2,769} \\
		\bottomrule
	\end{tabular}
\end{table}

\begin{table}[ht]
	\centering
	\caption{Class distribution across datasets (total instances, average per table).}
	\label{tab:datasets_stat}
	\setlength{\tabcolsep}{8pt}
	\begin{tabular}{lrrrrc}
		\toprule
		Dataset & Header & Column & Row & Section & Elem./table \\
		\midrule
		PubTables-1M & 876,027 & 4,924,504 & 10,145,446 & 651,749 & 18.9 \\
		Business & 4,362 & 38,457 & 30,326 & 2,737 & 17.4 \\
		Finance & 2,769 & 18,204 & 11,588 & 770 & 12.0 \\
		\bottomrule
	\end{tabular}
\end{table}

\section{Experimental Setting}
\subsection{Datasets}
We evaluate on a large-scale public benchmark and two private datasets. Statistics are summarized in Table~\ref{tab:datasets} and class distributions in Table~\ref{tab:datasets_stat}.
ConRTF detects four element classes that together define table structure. \emph{Rows} and \emph{columns} delineate the primary grid. \emph{Headers} are the top row(s) containing column labels, detected as a distinct class from data rows. \emph{Section separators} are full-width rows that span all columns, serving as group headings that partition the table into logical blocks. Data rows following a section separator are typically associated with it, though grouping conventions vary across layouts. For EFL supervision (Section~\ref{sec:efl}), headers, rows, and section separators are treated as row-like elements, while columns form the column-like class.

\textbf{PubTables-1M:} We adopt PubTables-1M as the primary public benchmark for Table Structure Recognition. We apply dataset-specific preprocessing to ensure annotation quality. Top-level headers are grouped and treated as a single header. Tables with excessive overlap (IoU >= 0.3) of between row-like elements or between columns are excluded, along with tables smaller than 200$\times$50 pixels or containing fewer than three structural elements. After filtering, the dataset contains 701,275 tables for training, 87,835 tables for validation, and 86,917 tables for testing. Since all methods are trained and evaluated on the same filtered splits, the results are directly comparable.

\textbf{Business Dataset}, consisting of 4,362 tables, with 654 tables used for testing. This dataset includes complex business tables with dense layouts and irregular structures, posing additional challenges for TSR.

\textbf{Finance Dataset} contains 2,769 tables, with 272 tables reserved for testing. The tables are extracted from structured financial documents and exhibit diverse header hierarchies and spanning patterns.

The use of in-house datasets alongside public benchmarks is established practice in TSR research: works such as TSRFormer~\cite{ACMMM_Lin_TSRFormer_2022} and RobusTabNet~\cite{PR_Ma_RobusTabNet_2023} validate on proprietary collections to test robustness under real-world conditions. PubTables-1M serves here as a fully reproducible public anchor, and the Business and Finance sets add domain diversity that public benchmarks alone do not cover.

\subsection{Evaluation Metrics}

We evaluate table structure recognition using two complementary metrics.

Tree Edit Distance-based Similarity (TEDS)~\cite{ECCV_Zhong_TEDS_2020} measures the minimum edit distance between HTML tree representations of predicted and ground-truth tables, normalized as $\mathrm{TEDS} = 1 - d / \max(|T_{\mathrm{pred}}|, |T_{\mathrm{GT}}|)$, where $d$ is the tree edit distance and $|T|$ denotes the number of nodes. TEDS\textsubscript{Struct} compares only structure (tags, row/column spans), while TEDS\textsubscript{Text} additionally evaluates cell content through character-level edit distance.

GriTS~\cite{ICDAR_Smock_GriTS_2023} evaluates table structure through three complementary dimensions based on optimal grid alignment. Ground-truth and predicted tables are represented as cell grids, and the Factored 2D Most Similar Substructures algorithm aligns them along rows and columns via dynamic programming. Pairwise cell similarities are then aggregated into precision, recall, and F-score. The three sub-metrics differ in their similarity function: GriTS\textsubscript{Top} compares cell topology (relative spans) via IoU, GriTS\textsubscript{Loc} compares cell bounding boxes via IoU, and GriTS\textsubscript{Con} compares cell text via longest common subsequence similarity. We report F-scores for all three sub-metrics.

\subsection{Implementation and Training Settings}
\label{sec:impl}

ConRTF adopts the D-FINE-X architecture~\cite{ICLR_Peng_DFINE_2025}: an HGNetv2-X backbone encoder~\cite{AX_Cui_HGNet_2021}, followed by a 6-layer Transformer decoder with 300 object queries. Input images are resized to $640 \times 640$. We optimize with AdamW using a learning rate of $2.5 \times 10^{-4}$ ($100\times$ smaller for the backbone), weight decay of $1.25 \times 10^{-4}$, and gradient clipping at $0.1$. Training uses exponential moving average (decay $0.9999$) and mixed-precision. Data augmentation includes photometric distortion, random zoom-out, IoU-based cropping, and horizontal flipping. We train for 20 epochs on PubTables-1M and up to 500 epochs with early stopping (patience 10) on the private datasets. All experiments run on a single NVIDIA A6000 GPU.

For the EFL loss (Section~\ref{sec:efl}), we measure element aspect ratios from the training annotations. Rows exhibit $\mathrm{AR} \approx 44$, headers $\approx 35$, section separators $\approx 10$, and columns $\approx 2$. Applying Eq.~\eqref{eq:lambda_opt} per class, the optima range from $\lambda_h \approx 1.7$ for rows and headers to $\lambda_h \approx 1.2$ for columns. Because EFL assigns a single weight pair to all classes, we adopt the column-constrained optimum $\lambda_h = 1.2$, $\lambda_l = 0.8$: this is the highest $\lambda_h$ that does not degrade defining-edge supervision for the least elongated class. For rows and headers, this value is below their per-class optimum but still provides meaningful improvement over the uniform baseline ($\lambda_h = \lambda_l = 1.0$). The weighting function parameters $a$ and $c$ in Eq.~\eqref{eq:weighting} and all loss component weights follow D-FINE defaults. The D-FINE-X baseline uses the identical architecture and training protocol with standard FGL and uniform edge weights ($\lambda_h = \lambda_l = 1.0$), providing a controlled ablation that directly isolates the contribution of edge-constrained supervision. We compare against TSR-optimized RT-DETRv2-X~\cite{CVPR_Zhao_RT-DETR_2024,AX_Lv_RT-DETRv2_2024}, YOLOv10-X~\cite{NeurIPS_Wang_YOLOv10_2024}, YOLO11-X~\cite{Jocher_YOLO11_2024}, and YOLO12-X~\cite{NeurIPS_Tian_YOLOv12_2025}.

\section{Experimental Results}
We evaluate ConRTF on PubTables-1M and two private datasets (Business and Finance). The experimental design centers on isolating the effect of edge-constrained supervision: ConRTF extends D-FINE-X with EFL while preserving the identical architecture, training protocol, and hyperparameters, enabling a direct measurement of EFL's contribution. Additional real-time baselines (RT-DETRv2, YOLOv10/11/12) provide broader context within the same speed class. We also analyze per-document score distributions and inference cost.

\subsection{Performance Comparison on PubTables-1M}

\begin{table}[ht]
\centering
\caption{Performance comparison on PubTables-1M. ConRTF improves over D-FINE-X on all metrics despite both operating above 99\%.}
\label{tab:sota_pubtables-1m}
\setlength{\tabcolsep}{6pt}
\begin{tabular}{llccccc}
\toprule
\multirow{2}{*}{Method} & \multirow{2}{*}{Dataset} &
\multicolumn{2}{@{}c@{}}{TEDS} & 
\multicolumn{3}{@{}c@{}}{GriTS} \\
\cmidrule(lr){3-4} \cmidrule{5-7}
& & Text & Struct & Top & Con & Loc \\
\midrule

YOLOv10-X & PubTables-1M & 96.50 & 96.79 & 98.64 & 92.57 & 96.35 \\ 
YOLO11-X & PubTables-1M & 96.83 & 97.14 & 96.64 & 90.92 & 94.70 \\ 
RT-DETRv2-X & PubTables-1M & 99.10 & 99.33 & 99.13 & 98.90 & 98.63 \\ 
D-FINE-X  & PubTables-1M & \underline{99.27} & \underline{99.40} & \underline{99.44} & \underline{99.29} & \underline{99.07} \\ 
Ours (ConRTF) & PubTables-1M & \textbf{99.36} & \textbf{99.47} & \textbf{99.51} & \textbf{99.37} & \textbf{99.18} \\

\bottomrule
\end{tabular}
\end{table}

Table~\ref{tab:sota_pubtables-1m} compares ConRTF with real-time baselines on PubTables-1M. A clear performance gap separates DETR-based architectures from YOLO-based detectors. RT-DETR, D-FINE, and ConRTF all exceed 99\% TEDS\textsubscript{Struct}, while YOLOv10-X and YOLO11-X remain around 97\%. The gap is most visible on GriTS\textsubscript{Con} (92.6 and 90.9 for the YOLOs vs.\ 99.3+ for DETR-based methods), indicating that imprecise boundary predictions affect content-level cell matching more than topology. ConRTF improves consistently over D-FINE across all metrics, confirming that EFL provides a measurable benefit even above 99\%. Beyond real-time detectors, ConRTF is competitive with sequence-based methods: its GriTS\textsubscript{Top} of 99.51 slightly exceeds the vision-language model UniTabNet (99.43)~\cite{EMNLP_Zhang_UniTabNet_2024}, a result it achieves as a single-pass detector without the token-by-token decoding that such methods require.

Aggregate scores above 99\% can obscure differences on the structurally complex tables that EFL targets. To quantify this, we isolate a hard subset of PubTables-1M where the D-FINE baseline scores below 99 GriTS\textsubscript{Con} ($n=4{,}322$, about 5\% of the test set). On this subset ConRTF reaches 88.4 GriTS\textsubscript{Con} against 85.8 for D-FINE, a gain of +2.6 points that is statistically significant ($p=1.5\times10^{-120}$). The same effect holds on the private datasets under the equivalent hard subset: +0.9 GriTS\textsubscript{Con} on Business ($n=294$, $p=0.005$) and +3.8 on Finance ($n=112$, $p=2\times10^{-4}$). The gains concentrate on tables with ambiguous or complex boundaries, where precise edge localization is hardest.

\subsection{Performance Under Limited Training Instances}

To evaluate generalization under limited training data, we test on two private datasets: Business (3,054 training and 654 test tables) and Finance (2,123 training and 272 test tables). In both cases, test documents were drawn from an independent pool, separate from the training source, ensuring no data leakage between splits. Results are reported in Table~\ref{tab_private_performance}.

\begin{table}[ht]
\centering
\setlength{\tabcolsep}{8pt}
\caption{Performance comparison on the private Business and Finance datasets. EFL gains increase with layout complexity, reaching +1.29 GriTS\textsubscript{Con} on Finance.}
\label{tab_private_performance}
\begin{tabular}{llccccc}
\toprule
\multirow{2}{*}{Method} & \multirow{2}{*}{Dataset} &
\multicolumn{2}{@{}c@{}}{TEDS} & 
\multicolumn{3}{@{}c@{}}{GriTS} \\
\cmidrule(lr){3-4} \cmidrule{5-7}
& & Text & Struct & Top & Con & Loc \\
\midrule

YOLOv10-X & Business & 68.09 & 71.43 & 81.33 & 68.18 & 66.60 \\ 
YOLO11-X & Business & 69.71 & 75.52 & 78.64 & 65.54 & 66.42 \\ 
YOLO12-X & Business & 64.61 & 71.88 & 75.20 & 62.03 & 63.38 \\ 
RT-DETRv2-X & Business & 88.51 & 91.03 & 89.41 & 87.18 & 85.18 \\ 
RT-DETRv2-L & Business & 89.49 & 91.73 & 88.17 & 86.00 & 84.12 \\ 
D-FINE-X  & Business & \underline{89.85} & \underline{92.01} & \underline{92.85} & \underline{90.96} & \underline{88.87} \\ 
Ours (ConRTF) & Business & \textbf{90.02} & \textbf{92.14} & \textbf{93.03} & \textbf{91.07} & \textbf{88.99} \\

\midrule

YOLOv10-X & Finance & 46.92 & 48.79 & 65.29 & 49.12 & 46.79 \\ 
YOLO11-X & Finance & 56.28 & 64.35 & 72.50 & 52.38 & 55.04 \\ 
YOLO12-X & Finance & 53.44 & 60.62 & 68.63 & 50.74 & 53.48 \\ 
RT-DETRv2-X & Finance & \underline{91.35} & \underline{92.97} & 91.88 & 90.20 & 87.79 \\ 
RT-DETRv2-L & Finance & 89.85 & 91.86 & 83.79 & 81.06 & 79.20 \\ 
D-FINE-X  & Finance & 91.18 & 92.65 & \underline{92.67} & \underline{91.12} & \underline{88.82} \\ 
Ours (ConRTF) & Finance & \textbf{91.98} & \textbf{93.67} & \textbf{94.28} & \textbf{92.41} & \textbf{89.49} \\

\bottomrule
\end{tabular}
\end{table}

With limited training data, the gap between DETR-based and YOLO-based methods widens dramatically. On Finance, YOLOv10-X drops to 48.8\% TEDS-\textsubscript{Struct}, a 48-point deficit relative to PubTables-1M, while DETR-based models maintain above 92\%. Notably, YOLO12-X's larger capacity does not improve performance on these smaller datasets, where it generally trails YOLO11-X. This underscores the value of geometric priors over raw capacity: ConRTF's gains come not from a larger model but from EFL's targeted boundary supervision.

ConRTF outperforms D-FINE on both datasets across all metrics. The gains are notably larger on Finance (+1.02 TEDS\textsubscript{Struct}, +1.29 GriTS\textsubscript{Con}) than Business (+0.13, +0.11), and the score distributions in Table~\ref{tab:score_dist} reveal why. Business tables are denser (17.4 elements per table vs.\ 12.0 for Finance, Table~\ref{tab:datasets_stat}) with more section separators, producing more catastrophic structural failures: 10.7\% of documents fall below 70\% GriTS\textsubscript{Con}, where the predicted structure fundamentally diverges from the ground truth. Finance has fewer such failures (6.6\% below 70\%) but a larger proportion of documents in the moderate band (21.3\% vs.\ 16.7\%), where errors stem from boundary imprecision rather than structural collapse. EFL directly addresses this through orientation-aware supervision, which is why the gains are largest on this dataset.

\subsection{Score Distribution Analysis}

Aggregate metrics can mask differences in per-document reliability. To complement the mean scores reported above, we analyze the distribution of per-document GriTS\textsubscript{Con} scores across three performance bands: high ($\geq 90\%$), moderate ($70$--$90\%$), and low ($< 70\%$).

\begin{figure*}[!t]
\centering
\begin{subfigure}[b]{\textwidth}
    \centering
    \includegraphics[width=0.95\textwidth]{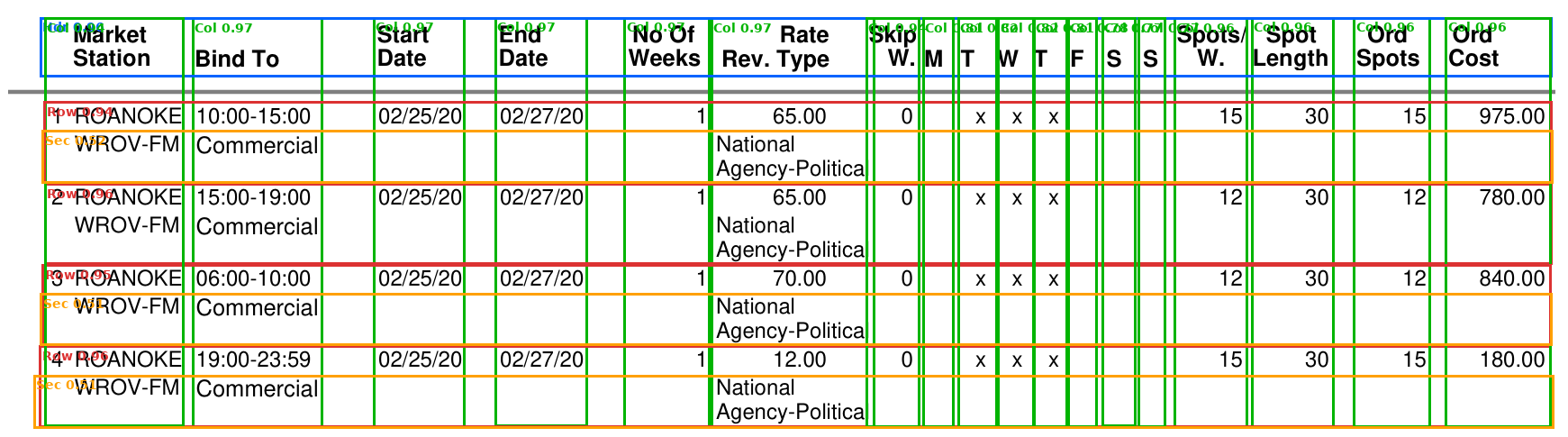}
    \caption{D-FINE-X (TEDS\textsubscript{Struct}: 0.37, GriTS\textsubscript{Con}: 0.33). Three spurious section detections (\textcolor{orange}{orange}) overlap with valid rows.}
    \label{fig:qual_doc1_dfine}
\end{subfigure}
\vspace{2pt}
\begin{subfigure}[b]{\textwidth}
    \centering
    \includegraphics[width=0.95\textwidth]{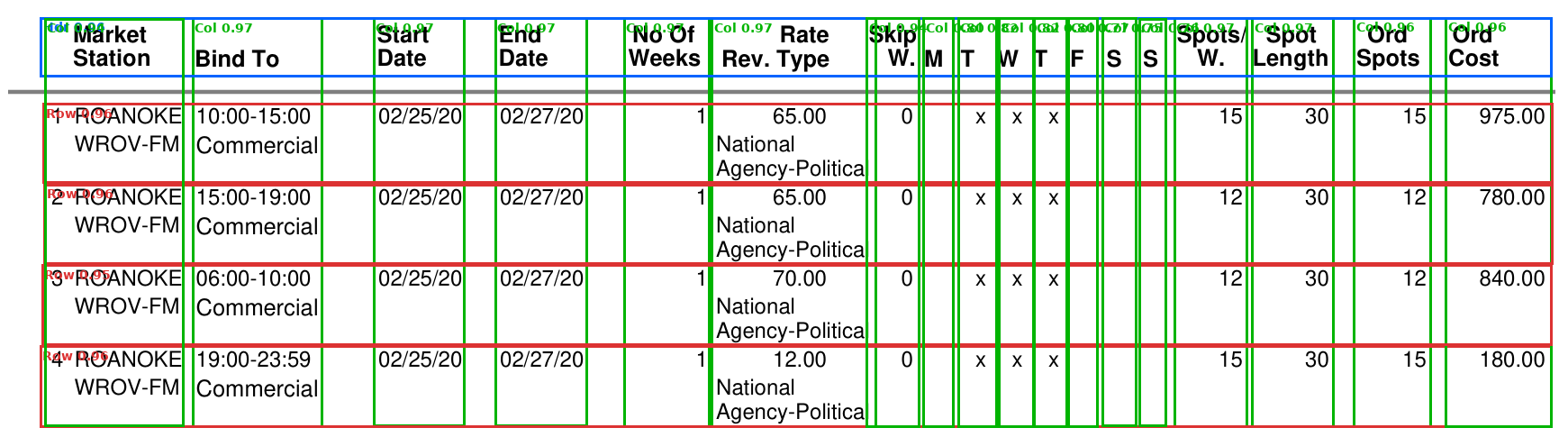}
    \caption{ConRTF (TEDS\textsubscript{Struct}: 0.84, GriTS\textsubscript{Con}: 0.98). No spurious sections. All four rows (\textcolor{red}{red}) are cleanly separated.}
    \label{fig:qual_doc1_conrtf}
\end{subfigure}
\vspace{4pt}
\noindent{\color{gray}\rule{\textwidth}{1.5pt}}
\vspace{4pt}
\begin{subfigure}[b]{\textwidth}
    \centering
    \includegraphics[width=0.95\textwidth]{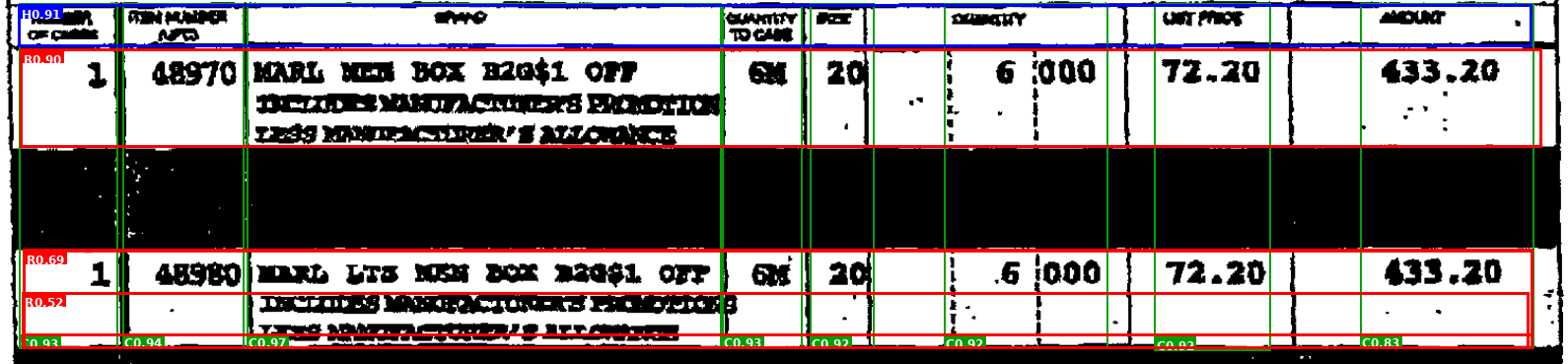}
    \caption{D-FINE-X (TEDS\textsubscript{Struct}: 0.75, GriTS\textsubscript{Con}: 0.75). Overlapping row boundaries at the bottom of the table produce duplicate row detections.}
    \label{fig:qual_doc2_dfine}
\end{subfigure}
\vspace{2pt}
\begin{subfigure}[b]{\textwidth}
    \centering
    \includegraphics[width=0.95\textwidth]{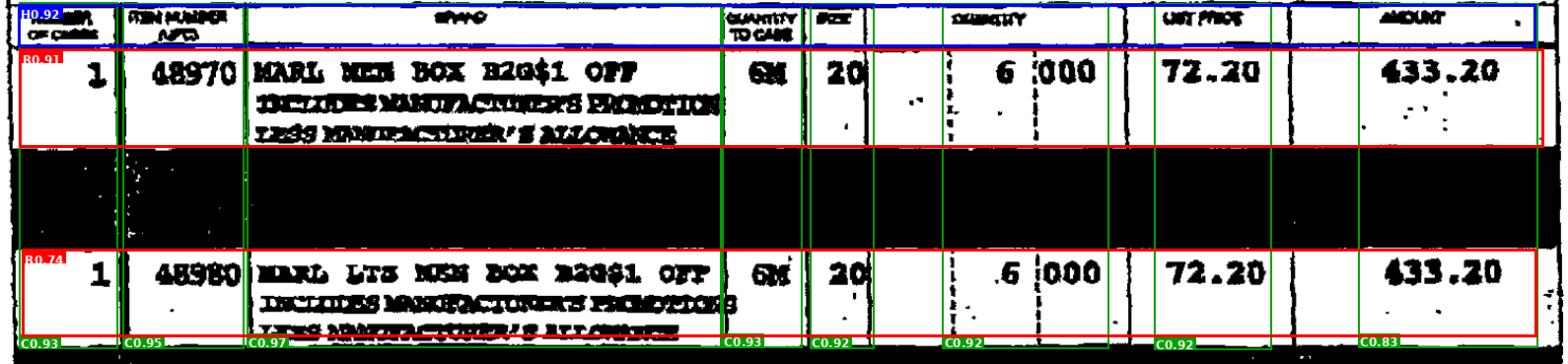}
    \caption{ConRTF (TEDS\textsubscript{Struct}: 1.0, GriTS\textsubscript{Con}: 1.0). EFL's edge-weighted supervision resolves the row boundary overlap, yielding exact structural recovery.}
    \label{fig:qual_doc2_conrtf}
\end{subfigure}
\caption{Qualitative comparison on two Business dataset tables. Colors: \textcolor{blue}{headers}, \textcolor{green}{columns}, \textcolor{red}{rows}, \textcolor{orange}{sections}. EFL supervision reduces spurious detections and boundary overlap, yielding cleaner structural predictions.}
\label{fig:qualitative}
\end{figure*}

Table~\ref{tab:score_dist} compares ConRTF with D-FINE, its architectural equivalent trained without EFL. On PubTables-1M, both models concentrate almost all documents in the high band, consistent with their near-identical aggregate scores. On the private datasets, where the training data is two to three orders of magnitude smaller, ConRTF consistently shifts documents toward higher scores. The effect is most pronounced on Finance: ConRTF places 72.1\% of documents above 90\% compared to 68.7\% for D-FINE ($+3.4$ pp), while reducing the proportion below 70\% from 10.7\% to 6.6\% ($-4.1$ pp). These gains indicate that EFL's orientation-aware supervision particularly benefits low-data regimes, where structural priors compensate for limited training signal (Figure~\ref{fig:qualitative}).

\begin{table}[ht]
\centering
\caption{Distribution of per-document GriTS\textsubscript{Con} scores (\% of test documents in each band). ConRTF vs.\ D-FINE (identical architecture, EFL loss only).}
\label{tab:score_dist}
\setlength{\tabcolsep}{10pt}
\begin{tabular}{llrrr}
\toprule
Dataset & Method & $\geq 90\%$ & $70$--$90\%$ & $< 70\%$ \\
\midrule
\multirow{2}{*}{PubTables-1M}
  & D-FINE  & 97.9 & 1.4 & 0.7 \\
  & ConRTF  & \textbf{98.2} & 1.2 & \textbf{0.6} \\
\midrule
\multirow{2}{*}{Business}
  & D-FINE  & 72.5 & 16.5 & 11.0 \\
  & ConRTF  & \textbf{72.6} & 16.7 & \textbf{10.7} \\
\midrule
\multirow{2}{*}{Finance}
  & D-FINE  & 68.7 & 20.6 & 10.7 \\
  & ConRTF  & \textbf{72.1} & 21.3 & \textbf{6.6} \\
\bottomrule
\end{tabular}
\end{table}

\subsection{Real-time Performance Analysis}

Table~\ref{tab:realtime} compares model complexity and inference speed across all methods. ConRTF and D-FINE-X share identical parameters (62M) and FLOPS (202G) because EFL operates only during training. The reported frames-per-second (FPS) values come from unoptimized PyTorch inference on an Intel Core i7-8850H CPU, a consistent setting for cross-method comparison: YOLOv10-X is fastest (2.0 FPS) with the fewest parameters (32M), while RT-DETRv2-X is slowest due to its larger architecture. On an NVIDIA RTX A6000 GPU, ConRTF reaches 15.42 FPS and matches D-FINE-X (14.61 FPS), which confirms that EFL adds no inference cost.

\begin{table}[H]
\centering
\caption{Inference cost comparison. FPS measured with unoptimized PyTorch on an Intel Core i7-8850H CPU. EFL operates only at training, so ConRTF and D-FINE-X are identical at inference.}
\label{tab:realtime}
\setlength{\tabcolsep}{10pt}
\begin{tabular}{lccc}
\toprule
\textbf{Method} & \textbf{Params} & \textbf{FLOPS} & \textbf{FPS (CPU)} \\
\midrule

YOLOv10-X & 32M & 171G & 2.0 \\
YOLO11-X & 57M & 196G & 1.6 \\
YOLO12-X & 59M & 200G & 1.4 \\
RT-DETRv2-X & 67M & 232G & 0.6 \\
RT-DETRv2-L & 32M & 108G & 1.0 \\
D-FINE-X  & 62M & 202G & 0.7 \\
Ours (ConRTF) & 62M & 202G & 0.7 \\

\bottomrule
\end{tabular}
\end{table}

Figure~\ref{fig:qualitative} illustrates two failure modes that EFL helps correct. In the first example (a--b), D-FINE produces three low-confidence section detections that overlap with valid row boundaries, fragmenting the table structure and dropping TEDS\textsubscript{Struct} to 0.37. ConRTF eliminates these spurious sections entirely, recovering all four data rows and raising TEDS\textsubscript{Struct} to 0.84. In the second example (c--d), D-FINE produces overlapping row boundaries at the bottom of the table, generating duplicate row detections that corrupt the grid structure. ConRTF resolves these boundary ambiguities, recovering the exact table layout with perfect TEDS and GriTS scores. In both cases, the errors involve boundaries between row-like elements, precisely the edges that EFL emphasizes through stronger horizontal supervision.

\section{Conclusion}
ConRTF reformulates table structure recognition by modeling table elements as row-like and column-like objects with distribution-based boundary representations. The Edge-constrained Fine-grained Localization (EFL) loss embeds orientation-aware geometric priors into boundary supervision, with edge weights derived from an IoU-sensitivity analysis. EFL operates during training only and adds no inference cost. Experiments on PubTables-1M and two private datasets show consistent improvements under TEDS and GriTS metrics, with the strongest gains in low-data regimes. The algorithm targets axis-aligned layouts. Extending EFL to per-class or instance-adaptive edge weights and to extremely low-data or zero-shot settings are promising directions for future work.

\begin{credits}
\subsection*{\ackname}
This research was supported by the CIFRE PhD program funded by the ANRT, by the European Union AI4DH project
(HORIZON-WIDERA-2023-TALENTS-01-01, grant 101186647), and by the YOOZ company. Views and opinions expressed are those of the author(s) only and do not necessarily reflect those of the European Union. Neither the European Union nor the granting authority can be held responsible for them.

\subsection*{\discintname}
The authors have no competing interests to declare that are relevant to the content
of this article.
\end{credits}

\bibliographystyle{splncs04}
\bibliography{refs}

@InProceedings{ICDAR_Hou_TABLET_2025,
	author="Hou, Qiyu and Wang, Jun",
	title="{TABLET}: Table Structure Recognition Using Encoder-only Transformers",
	booktitle="Document Analysis and Recognition (ICDAR)",
	year="2025",
	pages="253--278",
}

@article{PR_LONG_LOREPlus_2025,
	title = "{LORE++}: Logical location regression network for table structure recognition with pre-training",
	journal = {Pattern Recognition},
	volume = {157},
	pages = {110816},
	year = {2025},
	issn = {0031-3203},
	author = {Rujiao Long and Hangdi Xing and Zhibo Yang and Qi Zheng and Zhi Yu and Fei Huang and Cong Yao},
}

@article{ESA_Xiao_CascadeTSRDet_2025,
	title = {Rethinking detection based table structure recognition for visually rich document images},
	journal = {Expert Systems with Applications},
	volume = {269},
	pages = {126461},
	year = {2025},
	issn = {0957-4174},
	doi = {https://doi.org/10.1016/j.eswa.2025.126461},
	author = {Bin Xiao and Murat Simsek and Burak Kantarci and Ala Abu Alkheir}
}

@InProceedings{ICDAR_ThomasQUEST_2025,
  title="{QUEST}: Quality-aware Semi-supervised Table Extraction for Business Documents",
  author={Eliott Thomas and Mickael Coustaty and Aurélie Joseph and Gaspar Deloin and Elodie Carel and Vincent Poulain D'Andecy and Jean-Marc Ogier},
  booktitle="International Conference on Document Analysis and Recognition (ICDAR)",
  year={2025},
  doi={10.1007/978-3-032-04630-7_16},
}

@InProceedings{IJCAI_KhangTFLOP_2024,
  title="{TFLOP}: Table Structure Recognition Framework with Layout Pointer Mechanism",
  author={Minsoo Khang and Teakgyu Hong},
  booktitle="International Joint Conference on Artificial Intelligence (IJCAI)",
  year={2024}  
}

@InProceedings{EMNLP_Zhang_UniTabNet_2024,
  title="{UniTabNet}: Bridging Vision and Language Models for Enhanced Table Structure Recognition", 
  author={Zhenrong Zhang and Shuhang Liu and Pengfei Hu and Jiefeng Ma and Jun Du and Jianshu Zhang and Yu Hu},
  year={2024},
  booktitle="Findings of the Association for Computational Linguistics: EMNLP",
  pages = {6131–6143},
}

@InProceedings{CVPR_Smock_PubTables1M_2022,
    author    = {Smock, Brandon and Pesala, Rohith and Abraham, Robin},
    title     = "{PubTables-1M}: Towards Comprehensive Table Extraction From Unstructured Documents",
    booktitle = {IEEE/CVF Conference on Computer Vision and Pattern Recognition (CVPR)},
    year      = {2022},
    pages     = {4634-4642}
}

@InProceedings{ICDAR_Smock_TATR_2023,
    title     = {Aligning Benchmark Datasets for Table Structure Recognition},
    author    = {Smock, Brandon and Pesala, Rohith and Abraham, Robin},
    booktitle = {International Conference on Document Analysis and Recognition (ICDAR)},
    year      = {2023},
    pages	  ={371-386},
}

@InProceedings{CVPR_Zhao_RT-DETR_2024,
    author    = {Zhao, Yian and Lv, Wenyu and Xu, Shangliang and Wei, Jinman and Wang, Guanzhong and Dang, Qingqing and Liu, Yi and Chen, Jie},
    title     = "{DETR}s Beat {YOLO}s on Real-time Object Detection",
    booktitle = {IEEE/CVF Conference on Computer Vision and Pattern Recognition (CVPR)},
    year      = {2024},
    pages     = {16965-16974}
}

@misc{AX_Lv_RT-DETRv2_2024,
      title="{RT-DETRv2}: Improved Baseline with Bag-of-Freebies for Real-Time Detection Transformer", 
      author={Wenyu Lv and Yian Zhao and Qinyao Chang and Kui Huang and Guanzhong Wang and Yi Liu},
      year={2024},
      url={https://arxiv.org/abs/2407.17140}, 
}

@InProceedings{ICLR_Peng_DFINE_2025,
      title="{D-FINE}: Redefine Regression Task in {DETR}s as Fine-grained Distribution Refinement",
      author={Yansong Peng and Hebei Li and Peixi Wu and Yueyi Zhang and Xiaoyan Sun and Feng Wu},
      year={2025},
      booktitle = {International Conference on Learning Representations (ICLR)}
}

@inproceedings{CVPR_Xiang_GFocalv2_2021,
	author={Li, Xiang and Wang, Wenhai and Hu, Xiaolin and Li, Jun and Tang, Jinhui and Yang, Jian},
	title = {Generalized Focal Loss V2: Learning Reliable Localization Quality Estimation for Dense Object Detection},
	year = {2021},
	booktitle = {IEEE/CVF Conference on Computer Vision and Pattern Recognition (CVPR)},
}

@inproceedings{NeurIPS_Xiang_GFocal_2020,
	author = {Li, Xiang and Wang, Wenhai and Wu, Lijun and Chen, Shuo and Hu, Xiaolin and Li, Jun and Tang, Jinhui and Yang, Jian},
	title = {Generalized Focal Loss: Learning Qualified and Distributed Bounding Boxes for Dense Object Detection},
	year = {2020},
	booktitle = {International Conference on Neural Information Processing Systems (NeurIPS)},
}

@InProceedings{ECCV_Carion_DETR_2020,
	author="Carion, Nicolas and Massa, Francisco and Synnaeve, Gabriel and Usunier, Nicolas and Kirillov, Alexander and Zagoruyko, Sergey",
	title="End-to-End Object Detection with Transformers",
	booktitle="European Conference on Computer Vision (ECCV)",
	year="2020",
	pages="213--229",
}

@InProceedings{ECCV_Zhong_TEDS_2020,
	author="Zhong, Xu and ShafieiBavani, Elaheh and Jimeno Yepes, Antonio",
	title="Image-Based Table Recognition: Data, Model, and Evaluation",
	booktitle="European Conference on Computer Vision (ECCV)",
	year="2020",
	pages="564--580",
}

@InProceedings{ICDAR_Smock_GriTS_2023,
	author = {Smock, Brandon and Pesala, Rohith and Abraham, Robin},
	title="{GriTS}: Grid Table Similarity Metric for Table Structure Recognition",
	booktitle="Document Analysis and Recognition (ICDAR)",
	year="2023",
	pages = {535–549},
}

@InProceedings{WACV_Eliott_RAPTOR_2025,
	author={Eliott Thomas and Mickael Coustaty and Aurelie Joseph and Gaspar Deloin and Elodie Carel and Vincent Poulain D'Andecy and Jean-Marc Ogier},
	title="{RAPTOR}: Refined Approach for Product Table Object Recognition",
	booktitle="Winter Conference on Applications of Computer Vision (WACV)",
	year="2025"
}

@article{AX_Xiao_TableCenterNet_2025,
  title={Towards One-Stage End-to-End Table Structure Recognition with Parallel Regression for Diverse Scenarios},
  author={Anyi Xiao and Cihui Yang},
  journal={ArXiv},
  year={2025},
  volume={abs/2504.17522}
}

@InProceedings{ICDAR_Maksym_TSR_2023,
	title={Optimized Table Tokenization for Table Structure Recognition},
	author={Maksym Lysak and Ahmed Nassar and Nikolaos Livathinos and Christoph Auer and Peter W. J. Staar},
	booktitle="Document Analysis and Recognition (ICDAR)",
	year="2023",
}

@inproceedings{NeurIPS_Wang_YOLOv10_2024,
  title={YOLOv10: Real-Time End-to-End Object Detection},
  author={Wang, Ao and Chen, Hui and Liu, Lihao and Chen, Kai and Lin, Zijia and Han, Jungong and Ding, Guiguang},
  booktitle={Advances in Neural Information Processing Systems (NeurIPS)},
  year={2024},
}

@software{Jocher_YOLO11_2024,
  title={Ultralytics {YOLO11}},
  author={Jocher, Glenn and Qiu, Jing},
  year={2024},
  url={https://github.com/ultralytics/ultralytics},
  license={AGPL-3.0}
}

@inproceedings{NeurIPS_Tian_YOLOv12_2025,
  title={YOLOv12: Attention-Centric Real-Time Object Detectors},
  author={Tian, Yunjie and Ye, Qixiang and Doermann, David},
  booktitle={Advances in Neural Information Processing Systems (NeurIPS)},
  year={2025},
}

@article{AX_Cui_HGNet_2021,
  title={Beyond Self-Supervision: A Simple Yet Effective Network Distillation Alternative to Improve Backbones},
  author={Cui, Cheng and Guo, Ruoyu and Du, Yuning and He, Dongliang and Li, Fu and Wu, Zewu and Liu, Qiwen and Wen, Shilei and Huang, Jizhou and Hu, Xiaoguang and Yu, Dianhai and Ding, Errui and Ma, Yanjun},
  journal={arXiv preprint arXiv:2103.05959},
  year={2021},
}

@article{IJDAR_Zanibbi_Survey_2004,
  title={A Survey of Table Recognition: Models, Observations, Transformations, and Inferences},
  author={Zanibbi, Richard and Blostein, Dorothea and Cordy, James R.},
  journal={International Journal on Document Analysis and Recognition (IJDAR)},
  volume={7},
  number={1},
  pages={1--16},
  year={2004},
}

@article{ACM_Kasem_Survey_2024,
  title={Deep Learning for Table Detection and Structure Recognition: A Survey},
  author={Kasem, Mahmoud and Abdallah, Abdelrahman and Berendeyev, Alexander and Elkady, Ebrahem and Mahmoud, Mohamed and Abdalla, Mahmoud and Hamada, Mohamed and Nurseitov, Daniyar and Taj-Eddin, Islam},
  journal={ACM Computing Surveys},
  volume={56},
  number={12},
  year={2024},
}

@inproceedings{ICDAR_Schreiber_DeepDeSRT_2017,
  title={Deep{DeSRT}: Deep Learning for Detection and Structure Recognition of Tables in Document Images},
  author={Schreiber, Sebastian and Agne, Stefan and Wolf, Ivo and Dengel, Andreas and Ahmed, Sheraz},
  booktitle={International Conference on Document Analysis and Recognition (ICDAR)},
  pages={1162--1167},
  year={2017},
}

@inproceedings{CVPR_Nassar_TableFormer_2022,
  title={{TableFormer}: Table Structure Understanding with Transformers},
  author={Nassar, Ahmed and Livathinos, Nikolaos and Lysak, Maksym and Staar, Peter},
  booktitle={IEEE/CVF Conference on Computer Vision and Pattern Recognition (CVPR)},
  pages={4614--4623},
  year={2022},
}

@inproceedings{ICLR_Zhu_DeformableDETR_2021,
  title={Deformable {DETR}: Deformable Transformers for End-to-End Object Detection},
  author={Zhu, Xizhou and Su, Weijie and Lu, Lewei and Li, Bin and Wang, Xiaogang and Dai, Jifeng},
  booktitle={International Conference on Learning Representations (ICLR)},
  year={2021},
}

@inproceedings{ICLR_Zhang_DINO_2023,
  title={{DINO}: {DETR} with Improved DeNoising Anchor Boxes for End-to-End Object Detection},
  author={Zhang, Hao and Li, Feng and Liu, Shilong and Zhang, Lei and Su, Hang and Zhu, Jun and Ni, Lionel M. and Shum, Heung-Yeung},
  booktitle={International Conference on Learning Representations (ICLR)},
  year={2023},
}

@article{AIR_Sapkota_YOLO_Survey_2025,
  title={{YOLO} Advances to Its Genesis: A Decadal and Comprehensive Review of the You Only Look Once Series},
  author={Sapkota, Ranjan and Flores-Calero, Marco and Qureshi, Rizwan and others},
  journal={Artificial Intelligence Review},
  volume={58},
  number={9},
  pages={274},
  year={2025},
}

@inproceedings{ICML_Yang_GWD_2021,
  title={Rethinking Rotated Object Detection with Gaussian Wasserstein Distance Loss},
  author={Yang, Xue and Yan, Junchi and Ming, Qi and Wang, Wentao and Zhang, Xiaopeng and Tian, Qi},
  booktitle={International Conference on Machine Learning (ICML)},
  year={2021},
}

@inproceedings{NeurIPS_Yang_KLD_2021,
  title={Learning High-Precision Bounding Box for Rotated Object Detection via Kullback-Leibler Divergence},
  author={Yang, Xue and Yang, Xiaojiang and Yang, Jirui and Ming, Qi and Wang, Wentao and Tian, Qi and Yan, Junchi},
  booktitle={Advances in Neural Information Processing Systems (NeurIPS)},
  year={2021},
}

@article{TGRS_Zeng_ARSDETR_2024,
  title={{ARS-DETR}: Aspect Ratio-Sensitive Detection Transformer for Aerial Oriented Object Detection},
  author={Zeng, Ying and Yang, Xue and others},
  journal={IEEE Transactions on Geoscience and Remote Sensing},
  volume={62},
  year={2024},
}

@InProceedings{ACMMM_Lin_TSRFormer_2022,
  title="{TSRFormer}: Table Structure Recognition with Transformers",
  author={Weihong Lin and Zheng Sun and Chixiang Ma and Mingze Li and Jiawei Wang and Lei Sun and Qiang Huo},
  booktitle={Proceedings of the 30th ACM International Conference on Multimedia},
  year={2022},
}

@article{PR_Ma_RobusTabNet_2023,
  title={Robust Table Detection and Structure Recognition from Heterogeneous Document Images},
  author={Chixiang Ma and Weihong Lin and Lei Sun and Qiang Huo},
  journal={Pattern Recognition},
  volume={133},
  pages={109006},
  year={2023},
  issn={0031-3203},
}

\end{document}